\documentclass{article}
\usepackage{ijcai22}

\usepackage{times}
\usepackage{soul}
\usepackage{url}
\usepackage[hidelinks]{hyperref}
\usepackage[utf8]{inputenc}
\usepackage[small]{caption}
\usepackage{graphicx}
\usepackage{amsmath}
\usepackage{amsthm}
\usepackage{booktabs}
\usepackage{algorithm}
\usepackage{algorithmic}
\usepackage{amssymb}
\usepackage{color}
\usepackage{enumitem}
\usepackage{multirow}
\usepackage{makecell}
\usepackage{mathtools}
\usepackage{dsfont}

\title{Online Hybrid Lightweight Representations Learning: \\ Its Application to Visual Tracking}

\author{
\hspace{5mm}
Ilchae Jung$^{1,4}$ \hspace{15mm}
Minji Kim$^{1,2}$ \hspace{15mm}
Eunhyeok Park$^{4}$ \hspace{15mm}
Bohyung Han$^{1,2,3}$
\emails
\hspace{2mm}
\texttt{\small chey0313@postech.ac.kr} \hspace{3mm}
\texttt{\small minji@snu.ac.kr} \hspace{7mm}
\texttt{\small eh.park@postech.ac.kr} \hspace{8mm}
\texttt{\small bhhan@snu.ac.kr}  \hspace{10mm}
\affiliations
$^1$ASRI, $^2$ECE, $^3$IPAI, Seoul National University\\
$^4$CSE, POSTECH\\
}

\begin{document}

\maketitle

\begin{abstract}
This paper presents a novel hybrid representation learning framework for streaming data, where an image frame in a video is modeled by an ensemble of two distinct deep neural networks; one is a low-bit quantized network and the other is a lightweight full-precision network. 
The former learns coarse primary information with low cost while the latter conveys residual information for high fidelity to original representations.
The proposed parallel architecture is effective to maintain complementary information
since fixed-point arithmetic can be utilized in the quantized network and the lightweight model provides precise representations given by a compact channel-pruned network.
We incorporate the hybrid representation technique into an online visual tracking task, where deep neural networks need to handle temporal variations of target appearances in real-time. 
Compared to the state-of-the-art real-time trackers based on conventional deep neural networks, our tracking algorithm demonstrates competitive accuracy on the standard benchmarks with a small fraction of computational cost and memory footprint.
\end{abstract}

\section{Introduction}

Recent studies on deep neural representation learning~\cite{he2016deep,szegedy2015going} have achieved remarkable progress in various computer vision tasks.
However, since their success depends heavily on high-performance GPUs and large memory, it is still challenging to deploy trained models to real-world applications in resource-hungry environments, \textit{e.g.}, visual surveillance, robot navigation, and embedded systems.
Such drawbacks are aggravated in the problems handling videos, which critically demand efficient representation learning methodologies for real-time processing.

The mainstream approaches for reducing computational cost in neural networks are categorized into two groups: low-bit quantization and channel pruning.
Low-bit quantization methods~\cite{jung2019learning,bulat2019xnor} attempt to reduce the computational burden of deep neural networks by using low-precision weights and activations.
Although these approaches are successful under simple offline learning scenarios for the tasks in the image domain, \textit{e.g.}, image classification~\cite{russakovsky2015imagenet}, they still suffer from the challenges in streaming data including online adaptation and fine-grained prediction with limited precision.
The other paradigm of model compression, channel pruning focuses on reducing the number of channels in each layer of deep neural networks~\cite{he2017channel,gao2018dynamic,kang2020operation}.
%
Recently, a couple of channel pruning methods~\cite{jung2020real,che2018channel} have shown competitive performance to the baseline models with full parameters on online tasks for streaming data.
However, their compression efficiency concerning the size and inference cost remains inferior to the low-bit quantization methods.

By observing the complementary characteristics of the two methods, we propose a novel framework to learn hybrid representations for streaming data based on an ensemble of low-bit quantization and compact full-precision networks.
In the fusion of the hybrid networks, the low-bit quantized network delivers primary information of data efficiently while the lightweight channel-pruned network augments missing details in the representation and facilitates online model update to handle temporal variations of data.
We evaluate the proposed algorithm on online visual tracking, which requires online model updates and highly accurate bounding box alignments in real-time streaming data.
Our algorithm is unique in the sense that the output model is lightweight and suitable for systems with resource constraints.
Moreover, we empirically verify that our hybrid representation model is applicable to various types of tracking approaches~\cite{jung2018realtime,li2019siamrpn++,chen2021transformer}.
Our contributions are three-folded as follows: 
\begin{itemize}
\item We propose a hybrid representations learning framework based on low-bit quantization and channel pruning by learning an efficient low-bit-width primary network and a precise residual network, respectively.
\item We successfully deal with the trade-off between two popular model compression techniques and facilitate online adaptation of the hybrid network on streaming data by incorporating a residual architecture.
\item Our learning algorithm achieves competitive performance in terms of accuracy and efficiency and, in particular, illustrates competency in the resource-aware environment.
\end{itemize}

\section{Related Works}
\label{sec:related}

\subsection{Low-bit Quantization}
Recent studies on low-bit quantization~\cite{jung2019learning,park2020profit} have made successful progress in image classification.
The research related to binary quantization~\cite{hubara2016binarized,bulat2019xnor} shows the potential of low-precision optimization, achieving reasonable accuracy only with binarized activations and parameters.
Other approaches based on the 4- or 5-bit quantization~\cite{jung2019learning,park2020profit} propose more practical solutions, accomplishing lower or negligible accuracy loss while enabling efficient fixed-point computation. 

On the other hand, there exists another research trend that attempts to boost the performance of quantized models with assistance from full precision networks~\cite{zhuang2020training}.
Specifically, they train the quantization networks simultaneously with parameter-shared full precision networks to allow the relevant gradient flow for better accuracy.
However, the full precision network is abandoned for inference, which loses the full potency of the hybrid structure. 
We judiciously design the framework that utilizes the benefit of both full-precision and quantized networks for visual tracking. 

To mitigate the instability of training on network quantization,  \cite{qin2020forward} learns binary neural networks with an additional regularizer that maximizes the entropy of the distribution of quantized model parameters.
On the other hand, \cite{yu2020low} proposes a new regularizer to let the distribution of quantized model parameters mimic the uniform distribution.
Although those methods achieve competitive performance in offline problems such as image classification, their accuracy is degraded substantially during online updates, making the application to visual tracking not straightforward.
This paper introduces the newly designed normalization method considering the quantization interval to maintain the feature quality during online model updates. 

\subsection{Channel Pruning}
Another prominent deep neural network lightening method, channel pruning, aims to reduce the number of channels in model parameters.
The pioneering work based on the LASSO regularization~\cite{he2017channel} enforces the sparsity of the model while maintaining the accuracy in the image classification task.
\cite{gao2018dynamic} has proposed the feature boosting method by dynamically pruning channels at run-time using the channel saliency predictor.
Besides, \cite{jung2020real} adopts the conventional channel pruning method~\cite{he2017channel} to online object tracking, showing considerable performance gain.
While the power of channel pruned models is valid for both online and offline learning, their compression rates remain inferior to low-bit quantization.

Our hybrid representation learning algorithm aims to improve accuracy and efficiency by combining complementary representations from a low-bit quantization network and a lightweight full precision network.
While maintaining primary information in the quantization network obtained from rich offline pretraining, the full precision network compressed by channel pruning allows straightforward the online model update as well as accurate real-valued representation learning.
  

\section{Hybrid Representation Learning}
\label{sec:online}


\subsection{Overview}
\label{sub:overview}
Our goal is to enhance the quality of representations using an ensemble of two lightweight deep neural networks obtained from quantization and channel pruning.
The quantized network is responsible for feature encoding of the primary time-invariant information while the channel-pruned model covers high-precision time-varing residual information. 

Given an input image $A_0$, the standard convolutional neural network (CNN) learns a feature extractor $\mathtt{f}_\textrm{conv}(\cdot ; \cdot)$ and estimates a feature map $A_\textrm{conv}$ as follows:
\begin{equation} \label{formula:conv_layer}
    A_\textrm{conv} \equiv A_{N_l}  = \mathtt{f}_\textrm{conv} (A_{0}; \mathcal{W}_\textrm{conv}),
\end{equation}
where $\mathcal{W}_\textrm{conv} = \{ W_1, ..., W_{N_l}\}$ is a collection of convolutional filters throughout the $N_l$ layers.
In the $i^\text{th}$ intermediate layer, we apply a convolution with auxiliary operations and derive a feature map, which is given by
\begin{equation}
A_{i} = \sigma(A_{i-1} * W_{i} ),
\end{equation}
where $\sigma(\cdot)$ denotes a non-linear activation function and $*$ indicates the convolution operator. 
%
Contrary to the conventional CNNs, our framework utilizes two networks in parallel, a quantized network $\mathtt{f}^q (\cdot ; \cdot)$ and a channel-pruned network $\mathtt{f^p} (\cdot ; \cdot)$,
which generate outputs respectively as 
\begin{align}
    A^\textrm{q} = \mathtt{f}^q(A_0; \mathcal{W}^\textrm{q}), ~~~ \mathcal{W}^\textrm{q}=\{ W_1^\textrm{q},...,W_{N_l}^\textrm{q} \}, \\
    A^\textrm{p} = \mathtt{f}^p(A_0; \mathcal{W}^\textrm{p}),  ~~~ \mathcal{W}^\textrm{p}=\{ W_1^\textrm{p},...,W_{N_l}^\textrm{p} \}.
\end{align}
Then, the proposed hybrid representation, $A^\textrm{h}$, is given by the sum of two outputs as follows:
\begin{equation}
A^\textrm{h} = A^q + A^p = \mathtt{f}^\textrm{q}(A_0; \mathcal{W}^q) + \mathtt{f}^\textrm{p}(A_0; \mathcal{W}^p). 
\end{equation}
%
We can train prediction networks of arbitrary tasks followed by the feature extraction.
Note that the hybrid representation is also utilized in the prediction networks; channel pruning is applied to all layers of the prediction networks while low-bit quantization is partially applied to the activation quantization.

\subsection{Low-Bit Quantization of Primary Network}
\label{sub:low-bit}
To maximize the benefit of the reduced precision network, 
we quantize both model parameters 
and activations 
in all $N_l$ layers.
We adopt the quantization method proposed in \cite{jung2019learning}, and introduce the additional regularization loss terms to the quantized data for fast and stable training.

\subsubsection{Weight Quantization}

The weight quantization function $Q_W^l(\cdot): w_l \rightarrow w^q_l$ is composed of two steps.
First, the weights are linearly transformed to $[-1,1]$ by the learnable parameters $c^l_W$ and $d^l_W$, which define the center and the width of the quantization interval, respectively.
Only the weight values within the interval $[c_W^l-d_W^l, c_W^l+d_W^l]$ are quantized while the rests are clamped.
The formal definition of the first step is given by
\begin{equation}
\hat{w}_{l} =
\begin{cases}
\mathtt{sign}(w_{l}) & \text{if $|w_{l}| > c^l_{W} + d^l_{W}$ } \\
0 & \text{if $|w_{l}| < c^l_{W} - d^l_{W}$ } \\
\mathtt{sign}(w_{l}) \cdot (\alpha^l_{W} |w_{l}| + \beta^l_{W}) & \text{otherwise},
\end{cases}
\label{eq:quantization_w}
\end{equation}
where $\mathtt{sign}(\cdot) \in \{-1, 1\}$ means the sign of an input, $\alpha^l_{W} = 0.5/d^l_{W}$, and $\beta^l_{W}= -0.5 c^l_{W} / d^l_{W} + 0.5$. 
Second, the normalized weights are discretized as 
%
\begin{equation}
W^q_l = \mathtt{round}(\hat{W}_l \cdot (2^{N_b-1}-1)),
\label{eq:quantize_w}
\end{equation}
where $\mathtt{round}(\cdot)$ denotes the element-wise rounding and $N_b$ is the bit-width, leading to $(2^{N_b}-1)$ quantization levels. 

\subsubsection{Activation Quantization}
Similarly, the activation quantization function $Q_A^l(\cdot): a_l \rightarrow a^q_l$ has transformation and round steps.
First, the activations $a_l \in A_l$ in the $l^\text{th}$ layer are linearly transformed to the values in $[0, 1]$ as follows:
\begin{equation}
\hat{a}_l = 
\begin{cases}
1 & \text{if $a_l > c^l_{A} + d^l_{A}$ } \\
0 & \text{if $a_l < c^l_{A} - d^l_{A}$ } \\
\alpha^{l}_{A} A_l + \beta^{l}_{A} & \text{otherwise},
\end{cases}
\label{eq:quantization_a}
\end{equation}
where $\alpha^l_{A} = 0.5/d^l_{A}$ and $\beta^l_{A}= -0.5 c^l_{A} / d^l_{A} + 0.5$. 
The quantization interval is defined by the learnable parameters for its center ($c^l_A$) and width ($d^l_A$). 
We assume that the activations are non-negative due to the ReLU function.  
The final quantized output $A^q_l$ is obtained by applying the rounding function to the individual elements in $\hat{A}_l$,
%
\begin{equation}
A^q_l =  \mathtt{round}(\hat{A}_l \cdot (2^{N_b}-1)).
\label{eq:quantize}
\end{equation}
%

\subsubsection{Operation of Quantization Network}
The output from each convolutional layer is now given by $A_{l+1} = \sigma (Q_A^l(A_l) * Q_W^{l+1}(W_{l+1}) ),$
%
where the standard convolution operation is replaced by a hardware-friendly fixed-point operation for additional acceleration.
By applying the quantization to the whole network, the representation of the final convolutional layer is denoted by $A^q \equiv \mathtt{f}^q(A_0; \mathcal{W}^q)$,
%
where $\mathtt{f}^q (\cdot ; \cdot)$ is the feature extraction network via the quantization of both weights and activations.
%
%

\subsubsection{Training Quantization Networks with Normalization}
We now present our training scheme for robust and stable quantization. 
In particular, we discuss a novel regularization loss for quantization, $\mathcal{L}_\textrm{quant}$, which is divided into weight and activation normalization losses as 
\begin{align}
	\mathcal{L}_\textrm{quant} = \mathcal{L}_\textrm{qW} + \mathcal{L}_\textrm{qA}.
	\label{eq:loss_quant}
\end{align}

\paragraph{Weight normalization.}
Clipping model parameters using the learned interval $[c_W^l - d_W^l, c_W^l + d_W^l]$ leads to error in the quantized representations and makes the optimization procedure unstable.
It turns out that the final accuracy is sensitive to the initialization of $\{c^l_W, d^l_W\}$, especially with a lower bit-width.  
To alleviate the limitation, we propose a simple but effective regularizer, $\mathcal{L}_\textrm{qW}$, enforcing a majority of parameters to be located within a certain range, which is given by
%
\begin{align}
    \mathcal{L}_\textrm{qW} = \sum^{N_l}_{i=1} & \left(\mathtt{avg}({ \alpha^i_W |W_{i}| +\beta^i_W }) - \mu_0\right)^2 + \nonumber \\  
    & \left(\mathtt{std}({  \alpha^i_W |W_{i} | + \beta^i_W })-\sigma_0\right)^2 ,
\end{align}
%
where $\mathtt{avg}(\cdot)$ and $\mathtt{std}(\cdot)$ denote mean and standard deviation of all entries in their input matrices, respectively, $| \cdot |$ returns an absolute value matrix of the same size to its input, and $(\mu_0, \sigma_0)$ are hyperparameters for desirable mean and standard deviation values.
The proposed regularizer encourages the transformed model parameters to be in $[0,1]$ and reduces the error resulting from the clamping process.
Note that it allows the quantized network to identify more appropriate parameters $\{c^l_W, d^l_W\}$ and undergo more stable optimization.

\paragraph{Activation normalization.}
\label{paragraph:act_norm}
Clipping activations outside of the interval $[c^l_\textrm{A}-d^l_\textrm{A}, c^l_\textrm{A}+d^l_\textrm{A}]$ incurs critical information loss in practice.
To mitigate this issue, we define the following regularization loss term $\mathcal{L}_\textrm{qA}$:
%
\begin{align}
    \mathcal{L}_\textrm{qA} =  \sum^{N_l}_{i=1} &  \left[\mathtt{avg}(A_i^\textrm{bn}) + 2 \cdot \mathtt{std}(A_i^\textrm{bn}) - (c^l_\textrm{A}+d^l_\textrm{A}) \right]_{+} + \nonumber \\ 
    &  \left[ \sigma_i - \mathtt{std}(A_i^\textrm{bn}) \right]_{+}  + \left[ \mu_i - \mathtt{avg}(A_i^\textrm{bn}) \right]_{+},  
\end{align}
where $A^\textrm{bn}=\{  A_i^\textrm{bn}  \}_{i=1}^{N_l}$ is a set of activations after batch normalization and $\left[ \cdot \right]_+$ denotes the ReLU function.
Note that $\sigma_i$ and $\mu_i$ are determined by combination of current quantization interval parameters ($c^l_A, d^l_A$).
Assuming that the activation $A_i^\textrm{bn}$ follows a Gaussian distribution, the regularization loss enforces the activations larger than the mean of the Gaussian and lower than $\mathtt{avg}(A^\textrm{bn}_i)+2\cdot \mathtt{std}(A^\textrm{bn}_i)$ to be within the active range.
Statistically speaking, it implies that almost 96$\%$ of positive values are trained to be located in the active range.

\begin{figure*}
	\centering
		\includegraphics[width=0.9\linewidth]{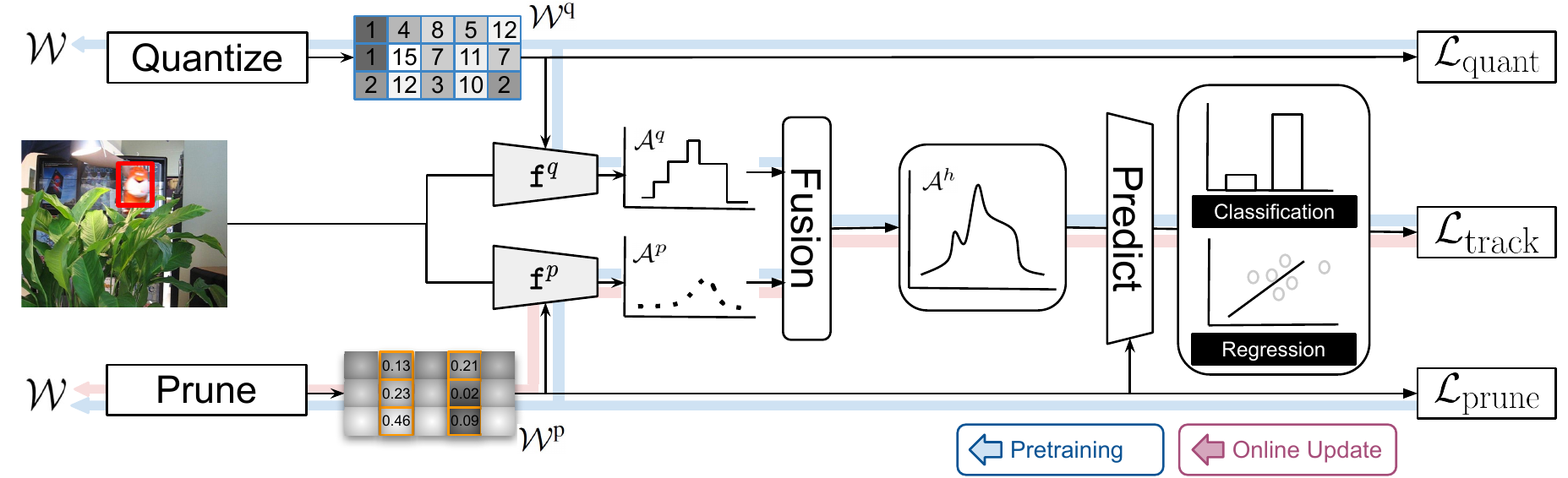}
	\caption{Overall framework of our algorithm. We train hybrid representation for both low-bit quantized network ($\mathtt{f}^q$) and lightweight full precision network ($\mathtt{f}^p$) on pretraining time while only the lightweight network is fine-tuned in test phase.
	Red arrow denotes the gradient backpropagation of full precision network in online model update while blue arrow denotes the backpropagation in pretraining. 
}
	\label{fig:overall_framework}
\end{figure*} 

\subsection{Channel Pruning of Residual Network}
\label{sub:channel}

We propose a stochastic channel pruning algorithm using a Gumbel-softmax trick to generate discrete mask samples from the model with a continuous output space.
Our algorithm learns a set of channel selection probability vectors, $\{ \mathbf{b}_1, ..., \mathbf{b}_{N_{l}}\}$, in each layer, where $\mathbf{b}_l$ indicates which channel is likely to be sampled in the $l^\textrm{th}$ layer.
By employing the Gumbel-softmax trick introduced in \cite{jang2016categorical}, $\mathbf{b}_l$ generates discrete channel selection mask $\mathbf{M}_{l}$ as follows:
\begin{align}
\label{formula:pruning}
\bar{\mathbf{M}}_l[i] & = \frac{\exp\left( \frac{g_i + \log (\mathbf{b}_l[i])}{\tau} \right) }{ \exp \left( \frac{g_i + \log (\mathbf{b}_l[i])}{\tau} \right) + \exp \left( \frac{g'_i + \log(1- \mathbf{b}_l[i])}{\tau} \right) }, \nonumber \\
\mathbf{M}_l[i] & = \mathtt{round}(\bar{\mathbf{M}}_l[i]) , 
\end{align}
where $g_i$ and $g'_i$ represent random noise samples from the Gumbel distribution, and $\tau$ denotes a temperature.
We apply the straight-through Gumbel estimator for gradient approximation in backpropagation similarly to \cite{jang2016categorical}.
In the forward pass of training, we apply the pruning mask $M_{l}$ to the output activation  $A^p_l$ in the $l^\textrm{th}$ layer, which is given by
\begin{equation}
A^p_{l} = \sigma ( A^p_{l-1} * W^p_{l} ) \odot \mathbf{M}_{l},  \quad l =1 \dots N_{l}
\end{equation}
where $\odot$ denotes an channel-wise multiplication.
Then, the representation of the final convolution layer is obtained by applying the channel pruning network $\mathtt{f}^p(\cdot ; \cdot)$, as follows: $A^p =  \mathtt{f}^p(A_0; \mathcal{W}^p)$. 
Note that a channel with a selection probability lower than the threshold is removed at testing.
For training, the channel selection probabilities $ \{ \mathbf{b}_l \}^{N_{l}}_{l=1}$ are learned by minimizing the following loss term 
\begin{align}
   \mathcal{L}_\textrm{prune} = \sum^{N_{l}}_{l=1} ||\bar{\mathbf{M}}_{l}||_1,
 \label{eq:loss_prune}
\end{align}
where the number of active channels is penalized.

\section{Hybrid Representation for Online Tracking}
\label{sec:visual}
We apply the proposed hybrid representation learning method to online visual tracking, which aims to sequentially estimate the location of a target object in a streamed video given the ground-truth bounding box at the first frame.
This task is appropriate for testing the proposed algorithm since visual tracking requires efficient online target appearance modeling and accurate bounding box regression.

The overall framework with online hybrid representation learning for visual tracking is illustrated in Figure~\ref{fig:overall_framework}, which presents two pathways of backpropagation for offline pretraining and online model updates.
In the pretraining stage, our framework trains the parameters of both the quantized network and the channel-pruned network.
On the other hand, during the online model update in test time, we freeze the parameters of the quantized network to maintain static coarse information while allowing the lightweight full precision network to adapt to target appearance changes.
We apply our algorithm on top of three baseline trackers, RT-MDNet~\cite{jung2018realtime}, SiamRPN++~\cite{li2019siamrpn++}, and TransT~\cite{chen2021transformer}.
We pretrain each tracker with a loss term 
\begin{align}
\mathcal{L}_\textrm{hyb} = \mathcal{L}_\textrm{baseline}+\lambda_1 \mathcal{L}_\textrm{quant}+\lambda_2 \mathcal{L}_\textrm{prune},
\end{align} 
where $\mathcal{L}_\textrm{baseline}$ denotes a loss term proposed from baseline tracker, and $\{\lambda_1,\lambda_2\}$ are the balancing hyperparmeters.

\paragraph{Hybrid RT-MDNet.}
RT-MDNet first generates a feature map from a whole image and then predicts candidate windows sampled from the feature map.
Hybrid RT-MDNet applies our hybrid representation network to the feature extraction network while the prediction network employs the channel-pruned network only.
We perform online updates with the channel-pruned network.

\paragraph{Hybrid SiamRPN++.}
SiamRPN++ deals with a tracking algorithm via the cross-correlation between the target representation and the candidate feature map using Siamese networks.
The predictor network, which is composed of a binary classifier and a bounding box regressor, estimates the confidence score of each adjusted candidate location based on the cross-correlation results.
To formulate Hybrid SiamRPN++,  we apply our hybrid representation network to the Siamese backbone and adopt channel-pruned network in the predictor.

\paragraph{Hybrid TransT.}
TransT is a variant of SiamRPN++ that applies the Transformer encoder-decoder networks~\cite{vaswani2017attention} to the feature correlation.
Similarly to Hybrid SiamRPN++,  we incorporate the hybrid network into the backbone and employ the channel-pruned network in other parts.
Additionally,  for all self-attention in the Transformer, we perform the activation quantization to reduce the computation burden in the dot-product operations.

\section{Experiments}
\label{sec:experiments}

\begin{table*}[htbp]
	\centering
	\scalebox{0.85}{%
	\renewcommand{\tabcolsep}{1.5mm}
	\begin{tabular}{l|rrrrrr|rrrrr|rrrrr}
		\toprule
		 & \multicolumn{6}{c|}{HybRT-MDNet} & \multicolumn{5}{c|}{HybSiamRPN++} & \multicolumn{5}{c}{HybTransT}\\
		 & conv1 & conv2 & conv3 & fc4 & fc5 & fc6 & $\ell$1 & $\ell$2 & $\ell$3 & $\ell$4 & RPN & $\ell$1 & $\ell$2 & $\ell$3 & att & pred \\
		\midrule
		Bit (A/W)  & 32/4 & 4/4  & 4/4 & - & -& - & 32/5& 5/5& 5/5 & 5/5 & 5/32 & 32/5& 5/5& 5/5 & 6/32 & 32/32 \\ 
		LayerRatio & 0.11 & 0.25 & 0.32 & 0.29 & 0.03 & 0.00 & 0.02& 0.04& 0.17& 0.51 & 0.26 & 0.02 & 0.06& 0.24& 0.65 & 0.03 \\ 
		$\text{RC}_\text{prune}$ & 0.19 & 0.04 & 0.09 & 0.53 & 0.26 & 0.49 & 0.09& 0.08& 0.07& 0.06& 1.00 & 0.07& 0.06& 0.06& 0.39 & 1.00 \\ 
		$\text{RC}_\text{quant}$ & 0.13 & 0.03 & 0.03 & - & - & - & 0.08& 0.04& 0.04& 0.04& - & 0.08& 0.04& 0.04&  0.16 & - \\ 
		$\text{RC}_\text{hybrid}$  & 0.32 & 0.07 & 0.12 & 0.53 & 0.26 & 0.49 & 0.17& 0.12& 0.12& 0.10& 1.00 & 0.15 & 0.10 & 0.11 &  0.55 &  1.00 \\
		\midrule
		$\text{RC}~(\text{BOPs}_\text{total}) $   & \multicolumn{6}{c|}{0.26 (= 4.9B / 19.0B)} & \multicolumn{5}{c|}{0.32 (= 5.1M / 15.2M)} & \multicolumn{5}{c}{0.42 (= 4.4M / 10.5M)} \\
		\bottomrule
	\end{tabular}}
\caption{The computational cost of the proposed approach.
We compare the relative amount of computation in channel-pruned networks, quantized networks, and our hybrid representation models compared to the original network.
Bit (A/W) denotes the bit-width of input activations (A) and parameters (W), and LayerRatio means the normalized computational overhead of each layer observed in the baseline network.
$\text{RC}_\text{prune}$, $\text{RC}_\text{quant}$, and $\text{RC}_\text{hybrid}$ indicate the relative cost of each layer compared to the baseline in the channel-pruned, quantized, and hybrid networks, respectively.
Finally, $\text{BOPs}_\text{total}$ presents the total BOPs of the proposed hybrid model and the baseline network.
	}
	\label{table:ablation_speed}	
\end{table*}
\begin{table}[t]
	\centering
	\scalebox{0.85}{
		\renewcommand{\tabcolsep}{2.5mm}
		\begin{tabular}{l|rr|rr}
			\toprule

			{} & {} & {} & \multicolumn{2}{c}{OTB2015}  \\
			Method     & Bit & $\text{RC} $ &  Prec & Succ  \\
			\midrule
			RT-MDNet    & 32 & 1.00 & \textbf{85.3} & 61.9  \\ 
			Q           & 4 & 0.20  & 82.0 &  58.7  \\
			Q + N  \eqref{eq:loss_quant}  & 4 & 0.20 & 83.7&  61.8 \\
			Q + N  \eqref{eq:loss_quant} + P \eqref{eq:loss_prune} & 4 & 0.26 & 84.9& \textbf{63.1}  \\
			\midrule
			SiamRPN++  & 32 & 1.00  & 90.5/87.6 & 69.5/66.3    \\
			Q             & 5  & 0.24 & 86.1 & 66.5 \\
			Q + N \eqref{eq:loss_quant}      & 5  & 0.24 & 87.3 & 67.1 \\
			Q + N \eqref{eq:loss_quant} + P \eqref{eq:loss_prune}  & 5  & 0.32 &  \textbf{89.5}  & \textbf{68.5}  \\
			\bottomrule
		\end{tabular}}
\caption{
Ablative experiment for training objectives.
Q, N, and P denote na\"ive quantization, normalization, and pruning, respectively.
In the OTB results of SiamRPN++, we present the scores reported in the paper and the reproduced accuracies.
}
		\label{table:ablation_objective}
\end{table}
\begin{table}[t]
	\centering
	\scalebox{0.85}{
		\renewcommand{\tabcolsep}{1.5mm}
	
		\begin{tabular}{l|rrrrrrr|rr}
			\toprule
			& \multicolumn{7}{c|}{Relative cost ($\text{RC}$) } & \multicolumn{2}{c}{OTB2015}  \\
			& conv1 & conv2 & conv3 & fc4 & fc5 & fc6 & total & Prec & Succ \\
			\midrule
			PQ  & 0.32 & 0.07 & 0.12 & 0.53 & 0.26 & 0.49 & 0.26 & 84.9 & \textbf{63.1}\\
			QQ  & 0.26 & 0.06 & 0.06 & 0.06 & 0.03 & 1.00 & 0.12  & 8.3 & 12.1 \\
			PP   & 0.80 & 0.32 & 0.78 & 0.58 & 0.35 & 0.58 & 0.58 & \textbf{86.6} & 61.8 \\
			\bottomrule
		\end{tabular}}
		\caption{Analysis of the combinations of the quantized network and the channel-pruned network in HybRT-MDNet.
We compare our hybrid representation (PQ) to Quantization-only ensemble representation (QQ) and Pruning-only ensemble representation (PP). 
}
		\label{table:qpqqpp}
\end{table}
\begin{table}[t]
	\centering \hspace{-0.3cm}
	\scalebox{0.85}{
		\renewcommand{\tabcolsep}{1.5mm}
		\begin{tabular}{l|rrrrrrr|rr}
			\toprule
			& \multicolumn{7}{c|}{Relative cost ($\text{RC}_\text{prune}$)} & \multicolumn{2}{c}{OTB2015} \\
			& conv1 & conv2 & conv3 & fc4 & fc5 & fc6 & total & Prec & Succ\\
			\midrule
			1 & 0.19 & 0.04 & 0.09 & 0.53 & 0.26 & 0.49& \textbf{0.22} & 84.9 & \textbf{63.1}\\ 
			2  & 0.32 & 0.10 & 0.10 & 0.53 & 0.26 & 0.51& 0.25& 85.2& 62.5 \\
			3  & 0.41 & 0.16 & 0.17 & 0.53 & 0.26 & 0.51& 0.30& 85.3 & 62.5 \\
			4  & 0.51 & 0.24 & 0.27 & 0.53 & 0.31 & 0.52& 0.37& \textbf{87.1} & 62.7 \\
			5  & 0.62 & 0.35 & 0.35 & 0.63 & 0.36 & 0.55& 0.46& 86.7 & 62.4 \\
			6  & 0.72 & 0.50 & 0.51 & 0.69 & 0.46 & 0.62& 0.58& 86.6 & 62.7 \\
			\bottomrule
		\end{tabular}}
		\caption{Accuracy-computation trade-off given by varying channel-pruning ratios in HybRT-MDNet.
	}
	\label{table:prune}
\end{table}
\begin{table}[t]	
	\centering
	\scalebox{0.85}{
		\renewcommand{\tabcolsep}{4.0mm}
		\begin{tabular}{c|rrrr }
			\toprule
			Bit-width & 2bit & 3bit & 4bit & 5bit  \\
			\midrule
			Prec  & 84.7 & 84.3  & 84.9 & \textbf{85.5}  \\ 
			Succ  & 59.9 & 60.7 & 63.1 & \textbf{63.2}  \\ 
			\bottomrule
		\end{tabular}}
\caption{Accuracy-computation trade-off given by varying bit-widths in HybRT-MDNet.}
	\label{table:quant}
\end{table}
\begin{table*}[t]
	\centering
	\scalebox{0.85}{
	\renewcommand{\tabcolsep}{2.0mm}
	\begin{tabular}{lrrrrrrrrrr}
		\toprule
		\multirow{2}{*}{Method} & \multirow{2}{*}{Bit} &
		\multirow{2}{*}{$\text{RC}$} & \multicolumn{2}{c}{OTB} & \multicolumn{2}{c}{UAV123} & \multicolumn{2}{c}{LaSOT}  & \multicolumn{1}{c}{VOT2016} & \multicolumn{1}{c}{VOT2018} \\ 
		\multirow{2}{*}{} & \multirow{2}{*}{}  & \multirow{2}{*}{} & Succ & Prec & Succ & Prec & Succ & Norm. Pr & EAO & EAO \\ 
		\midrule
		ATOM~\cite{danelljan2019atom}                 & 32  & - & 67.1 & 87.9 & 64.4  &  51.5    & 51.5 & 57.6 & - &  0.401 \\
		DaSiamRPN~\cite{zheng2018distractor}            & 32  & - & 65.8 & 88.0 & 58.6  & 79.6  & 41.5 & 49.6 &0.411 & 0.326 \\ 
		MDNet~\cite{nam2016learning}                  & 32  & - & 67.8 & 90.9 & 52.8  & 74.7  & 39.7 & 46.0 & 0.257 &  -    \\ 
		\midrule
		RT-MDNet~\cite{jung2018realtime}  & 32  &1.00 & 65.0 & 88.5& 52.8 & 77.2   & 30.8&  36.0 &  0.325  &  0.176 \\ 
		HybRT-MDNet    & 5  &  0.29  & 64.5 & 88.0& 52.6 &   79.5 & 29.5 & 35.1& 0.329  & 0.194 \\ 
		\midrule
	    SiamRPN++ \cite{li2019siamrpn++} & 32 &  1.00 & 69.5 & 90.5 & 60.6 &  79.8  & 49.3 & 57.7 & 0.408 & 0.340 \\ 
		SiamRPN++ (Rep) & 32 &  1.00 & 66.3 & 87.6 & 57.0 & 76.9 & 45.6 & 54.5 & 0.401 & 0.305 \\ 
		HybSiamRPN++     & 5 & 0.32 & 68.5 & 89.5 & 59.7& 78.8 & 45.9& 53.6 & 0.383 & 0.299   \\
		OnHybSiamRPN++ & 5 & 0.34 & 69.4  & 89.8 & 61.2 & 81.8 & 46.1 & 54.1 &  0.412 & 0.344   \\
		\midrule
		TransT~\cite{chen2021transformer} & 32 & 1.00 & 68.2 & 88.3 & 66.0 & 85.2 & 64.2 & 73.5 &  0.387 & 0.298\\
		HybTransT & 5 & 0.42 & 68.2 & 88.4 & 67.1& 86.5& 64.5 & 73.9 & 0.418 & 0.285 \\
		\bottomrule
		
	\end{tabular}}

	\caption{Experiments on several benchmarks with a comparison of other competitive trackers. 
Our hybrid models (HybRT-MDNet, HybSiamRPN++, OnHybSiamRPN++, HybTransT) present competitive performance compared to other trackers while their computational costs are significantly reduced.
}
	\label{table:several_benchmarks}
\end{table*}


\subsection{Preparation for Training}
To pretrain the hybrid trackers, we employ the same training datasets as the baseline networks~\cite{jung2018realtime,li2019siamrpn++,chen2021transformer}.
We adopt the pretrained model of each baseline tracker for initialization and fine-tune them using the proposed loss function.
The learned hybrid models are tested on the standard visual tracking benchmarks, OTB2015~\cite{wu2015object}, UAV123~\cite{matthias2016benchmark}, LaSOT~\cite{fan2019lasot}, VOT2016~\cite{kristan2016vot}, and VOT2018~\cite{Kristan2018vot}, for comparisons with recently proposed competitors.
We evaluate algorithms using Precision rate (Prec) and Success rate (Succ) in OTB2015, Expected Average of Overlap (EAO) in VOT2016/2018, and Normalized Precision (Norm. Pr) in LaSOT.

\subsection{Implementation Details}

We employ the same learning rates as the baselines but reduce the rate of the quantized network to 1/10 of the original value in Hybrid SiamRPN++.
As in \cite{jung2019learning}, we use the full precision in the first convolution layer in Hybrid SiamRPN++ and Hybrid TransT.


For Hybrid RT-MDNet, we train the model with batch size 48 and run 120K iterations while optimizing model parameters and auxiliary parameters using SGD and Adam, respectively.
%
For Hybrid SiamRPN++, training is performed on 8 GPUs for 100K iterations with batch size 22, and the same optimization methods are employed as Hybrid RT-MDNet.
%
Hybrid TransT is trained for 400 epochs (1K iterations per epoch) using the AdamW optimizer with batch size 40, where the learning rate is decayed by a factor of 10 after 250 epochs.

Two variants of SiamRPN++ are tested, HybSiamRPN++ and OnHybSiamRPN++, which correspond to the models based on offline and online hybrid representation learning, respectively.
Specifically, during the online update stage, OnHybSiamRPN++ fine-tunes the last linear layer in the classification branch stemmed from the first cross-correlation block every 10 frames for 15 iterations using the data collected from the latest 20 frames.
The sample features are collected after the target prediction and annotated by the same labeling strategy used in the pretraining stage.
Our algorithm is implemented in PyTorch with Intel I7-6850k and NVIDIA Titan Xp Pascal GPU for RT-MDNet and Quadro RTX4000 for SiamRPN++ and TransT.


\subsection{Effects on Acceleration}
Following~\cite{baskin2018uniq}, we measure the number of bit-wise operations (BOPs) for unit convolution to estimate the performance benefit of the low-precision computation. 
For each layer with $\{C^\textrm{in}, C^\textrm{out} \}$ input and output channels based on $K \times K$ kernels, BOPs is given by 
\begin{align}
\textrm{BOPs} = C^\textrm{in}  C^\textrm{out} K^2 (b_a b_w + b_a + b_w + \log_2{C^\textrm{in} K^2}),
\end{align}
where $b_a$ and $b_w$ are the precisions of input activations and parameters, respectively.
Table~\ref{table:ablation_speed} shows the layer-wise analysis of the effect of our optimization method.

The proposed hybrid representation learning approach reduces the computation overhead by maintaining a small fraction of parameters that are critical to accuracy.
Specifically, we observe more channels pruned in mid-layers (conv2-3 and $\ell$2-4) than the parameters in head layers (conv1 and $\ell$1) and the prediction layers (fc, RPN, att, and pred), which tend to be intact for the stability of training.
Consequently, the proposed model with hybrid representations only requires a fraction of BOPs (0.26 to 0.42 as presented in Table~\ref{table:ablation_speed}) compared to the original network.

\subsection{Effect of Hybrid Representation}
We analyze the impact of each component in the proposed hybrid representation learning.
Note that we use HybRT-MDNet pretrained in VOT-OTB dataset for the ablative study.

\paragraph{Benefit of each component.}
Table~\ref{table:ablation_objective} illustrates the benefit of each component in our hybrid representation.
The quantization with normalization (Q+N)~\eqref{eq:loss_quant} achieves better accuracy than the model by a simple quantization (Q), which implies the effectiveness of our normalization strategies for stable quantization.
In addition, the hybrid representation model (Q+N+P) trained with both \eqref{eq:loss_quant} and \eqref{eq:loss_prune} outperforms the Q+N model, demonstrating that the proposed hybrid representation learning improves performance effectively.
Finally, our final models present competitive accuracy to the baseline models while reducing the computation cost significantly.

\paragraph{Combinations of quantization and pruning.}
Table~\ref{table:qpqqpp} illustrates the performance resulting from several combinations of quantized and pruned networks, which include the proposed hybrid network (PQ), the ensemble of pruned networks (PP), and the ensemble of quantized networks (QQ). 
We perform this experiment on the RT-MDNet baseline.
The numbers in the table denote the relative computational cost in terms of BOPs between the model corresponding to each row and the RT-MDNet baseline.
According to the results, the proposed hybrid representation presents a competitive performance in terms of both accuracy and efficiency, whereas the ensemble of quantized networks and pruned networks suffer from performance degradation in accuracy and computational cost, respectively.

\paragraph{Impact of pruning rates.}
Table~\ref{table:prune} analyzes the computational cost by varying channel pruning ratios for our hybrid model. 
Note that our model based on the hybrid representation maintains competitive accuracy in both measures, precision and success rates, even after pruning almost 80\% of channels.

\paragraph{Impact of quantization bit-width.}
Table~\ref{table:quant} presents the experimental results by varying the bit-width for quantization in our hybrid model. 
The results show that the success rate drops sharply with less than 4-bit quantization while the precision rate is hardly affected by the bit-width variation.

\subsection{Comparison on multiple benchmarks}
Table~\ref{table:several_benchmarks} compares the performance of our hybrid networks with conventional models in visual tracking on the standard benchmarks, where
SiamRPN++ and SiamRPN++ (Rep) indicate the official model and its reproduced version employed in our experiment, respectively.
The proposed hybrid networks such as HybRT-MDNet, HybSiamRPN++, OnHybSiamRPN++, and HybTransT achieve competitive performance to the baseline models while significantly reducing the computational cost (0.29 to 0.42 in RC).
Besides, the modified Hybrid SiamRPN++ with online update (OnHybSiamRPN++) improves tracking accuracy compared to the baseline SiamRPN++, which implies the effectiveness of our online hybrid representation learning for streaming data. 
Our model with the hybrid representation based on Transformer, HybTransT, also accomplishes outstanding accuracy with the significantly reduced computational burden.

\section{Conclusion}
\label{sec:conclusion}
We presented a novel framework of online hybrid representation learning applicable to streaming data by an ensemble of an efficient low-bit quantized network and a precise lightweight full-precision network. 
Adopting a residual architecture, the online hybrid representation successfully complements the two model compression approaches.   
We further improved the performance of the networks based on the hybrid representation through effective normalization of weights and activations for low-bit quantization.
The proposed algorithm was incorporated into online visual tracking tasks and verified that the proposed hybrid representation makes competitive accuracy to baseline models despite low-bit quantization and channel pruning.



\section*{Acknowledgements}
This work was supported in part by Samsung Electronics Co., Ltd. (IO210917-08957-01), and by the IITP grants [2021-0-01343, Interdisciplinary Program in AI (SNU); 2021-0-02068, AI Innovation Hub] and the NRF grant [2021M3A9E4080782, 2022R1A2C3012210] funded by the Korean government (MSIT).

\clearpage

\bibliographystyle{named}
\bibliography{ref_test}

\begin{thebibliography}{}

\bibitem[\protect\citeauthoryear{Baskin \bgroup \em et al.\egroup
  }{2018}]{baskin2018uniq}
Chaim Baskin, Eli Schwartz, Evgenii Zheltonozhskii, Natan Liss, Raja Giryes,
  Alex~M Bronstein, and Avi Mendelson.
\newblock Uniq: Uniform noise injection for non-uniform quantization of neural
  networks.
\newblock {\em arXiv preprint arXiv:1804.10969}, 2018.

\bibitem[\protect\citeauthoryear{Bulat and Tzimiropoulos}{2019}]{bulat2019xnor}
Adrian Bulat and Georgios Tzimiropoulos.
\newblock Xnor-net++: Improved binary neural networks.
\newblock {\em arXiv preprint arXiv:1909.13863}, 2019.

\bibitem[\protect\citeauthoryear{Che \bgroup \em et al.\egroup
  }{2018}]{che2018channel}
Manqiang Che, Runling Wang, Yan Lu, Yan Li, Hui Zhi, and Changzhen Xiong.
\newblock Channel pruning for visual tracking.
\newblock In {\em ECCVW}, 2018.

\bibitem[\protect\citeauthoryear{Chen \bgroup \em et al.\egroup
  }{2021}]{chen2021transformer}
Xin Chen, Bin Yan, Jiawen Zhu, Dong Wang, Xiaoyun Yang, and Huchuan Lu.
\newblock Transformer tracking.
\newblock In {\em CVPR}, 2021.

\bibitem[\protect\citeauthoryear{Danelljan \bgroup \em et al.\egroup
  }{2019}]{danelljan2019atom}
Martin Danelljan, Goutam Bhat, Fahad~Shahbaz Khan, and Michael Felsberg.
\newblock Atom: Accurate tracking by overlap maximization.
\newblock In {\em CVPR}, 2019.

\bibitem[\protect\citeauthoryear{Fan \bgroup \em et al.\egroup
  }{2019}]{fan2019lasot}
Heng Fan, Liting Lin, Fan Yang, Peng Chu, Ge~Deng, Sijia Yu, Hexin Bai, Yong
  Xu, Chunyuan Liao, and Haibin Ling.
\newblock Lasot: A high-quality benchmark for large-scale single object
  tracking.
\newblock In {\em CVPR}, 2019.

\bibitem[\protect\citeauthoryear{Gao \bgroup \em et al.\egroup
  }{2018}]{gao2018dynamic}
Xitong Gao, Yiren Zhao, {\L}ukasz Dudziak, Robert Mullins, and Cheng-zhong Xu.
\newblock Dynamic channel pruning: Feature boosting and suppression.
\newblock {\em arXiv preprint arXiv:1810.05331}, 2018.

\bibitem[\protect\citeauthoryear{He \bgroup \em et al.\egroup
  }{2016}]{he2016deep}
Kaiming He, Xiangyu Zhang, Shaoqing Ren, and Jian Sun.
\newblock Deep residual learning for image recognition.
\newblock In {\em CVPR}, 2016.

\bibitem[\protect\citeauthoryear{He \bgroup \em et al.\egroup
  }{2017}]{he2017channel}
Yihui He, Xiangyu Zhang, and Jian Sun.
\newblock {Channel pruning for accelerating very deep neural networks}.
\newblock In {\em ICCV}, 2017.

\bibitem[\protect\citeauthoryear{Hubara \bgroup \em et al.\egroup
  }{2016}]{hubara2016binarized}
Itay Hubara, Matthieu Courbariaux, Daniel Soudry, Ran El-Yaniv, and Yoshua
  Bengio.
\newblock Binarized neural networks.
\newblock In {\em NeurIPS}, 2016.

\bibitem[\protect\citeauthoryear{Jang \bgroup \em et al.\egroup
  }{2017}]{jang2016categorical}
Eric Jang, Shixiang Gu, and Ben Poole.
\newblock Categorical reparameterization with gumbel-softmax.
\newblock In {\em ICLR}, 2017.

\bibitem[\protect\citeauthoryear{Jung \bgroup \em et al.\egroup
  }{2018}]{jung2018realtime}
Ilchae Jung, Jeany Son, Mooyeol Baek, and Bohyung Han.
\newblock {Real-Time MDNet}.
\newblock In {\em ECCV}, 2018.

\bibitem[\protect\citeauthoryear{Jung \bgroup \em et al.\egroup
  }{2019}]{jung2019learning}
Sangil Jung, Changyong Son, Seohyung Lee, Jinwoo Son, Jae-Joon Han, Youngjun
  Kwak, Sung~Ju Hwang, and Changkyu Choi.
\newblock Learning to quantize deep networks by optimizing quantization
  intervals with task loss.
\newblock In {\em CVPR}, 2019.

\bibitem[\protect\citeauthoryear{Jung \bgroup \em et al.\egroup
  }{2020}]{jung2020real}
Ilchae Jung, Kihyun You, Hyeonwoo Noh, Minsu Cho, and Bohyung Han.
\newblock Real-time object tracking via meta-learning: Efficient model
  adaptation and one-shot channel pruning.
\newblock {\em AAAI}, 2020.

\bibitem[\protect\citeauthoryear{Kang and Han}{2020}]{kang2020operation}
Minsoo Kang and Bohyung Han.
\newblock Operation-aware soft channel pruning using differentiable masks.
\newblock In {\em ICML}, 2020.

\bibitem[\protect\citeauthoryear{Kristan \bgroup \em et al.\egroup
  }{2016}]{kristan2016vot}
Matej Kristan, Ale\v{s} Leonardis, Jiri Matas, Michael Felsberg, Roman
  Pflugfelder, Luka \v{C}ehovin Zajc, Tomas Vojir, Gustav H\"{a}ger, Alan
  Luke\v{z}i\v{c}, and Gustavo Fernandez.
\newblock The visual object tracking {VOT2016} challenge results.
\newblock In {\em ECCVW}, 2016.

\bibitem[\protect\citeauthoryear{Kristan \bgroup \em et al.\egroup
  }{2018}]{Kristan2018vot}
Matej Kristan, Ales Leonardis, Jiri Matas, Michael Felsberg, Roman Pfugfelder,
  Luka \v{C}ehovin Zajc, Tomas Vojir, Goutam Bhat, Alan Lukezic, Abdelrahman
  Eldesokey, Gustavo Fernandez, and et~al.
\newblock The sixth visual object tracking vot2018 challenge results.
\newblock In {\em ECCVW}, 2018.

\bibitem[\protect\citeauthoryear{Li \bgroup \em et al.\egroup
  }{2019}]{li2019siamrpn++}
Bo~Li, Wei Wu, Qiang Wang, Fangyi Zhang, Junliang Xing, and Junjie Yan.
\newblock Siamrpn++: Evolution of siamese visual tracking with very deep
  networks.
\newblock In {\em CVPR}, 2019.

\bibitem[\protect\citeauthoryear{Mueller \bgroup \em et al.\egroup
  }{2016}]{matthias2016benchmark}
Matthias Mueller, Neil Smith, and Bernard Ghanem.
\newblock {A benchmark and simulator for UAV tracking}.
\newblock In {\em ECCV}, 2016.

\bibitem[\protect\citeauthoryear{Nam and Han}{2016}]{nam2016learning}
Hyeonseob Nam and Bohyung Han.
\newblock Learning multi-domain convolutional neural networks for visual
  tracking.
\newblock In {\em CVPR}, 2016.

\bibitem[\protect\citeauthoryear{Park and Yoo}{2020}]{park2020profit}
Eunhyeok Park and Sungjoo Yoo.
\newblock Profit: A novel training method for sub-4-bit mobilenet models.
\newblock In {\em ECCV}, 2020.

\bibitem[\protect\citeauthoryear{Qin \bgroup \em et al.\egroup
  }{2020}]{qin2020forward}
Haotong Qin, Ruihao Gong, Xianglong Liu, Mingzhu Shen, Ziran Wei, Fengwei Yu,
  and Jingkuan Song.
\newblock Forward and backward information retention for accurate binary neural
  networks.
\newblock In {\em CVPR}, 2020.

\bibitem[\protect\citeauthoryear{Russakovsky \bgroup \em et al.\egroup
  }{2015}]{russakovsky2015imagenet}
Olga Russakovsky, Jia Deng, Hao Su, Jonathan Krause, Sanjeev Satheesh, Sean Ma,
  Zhiheng Huang, Andrej Karpathy, Aditya Khosla, Michael Bernstein, et~al.
\newblock Imagenet large scale visual recognition challenge.
\newblock {\em IJCV}, 2015.

\bibitem[\protect\citeauthoryear{Szegedy \bgroup \em et al.\egroup
  }{2015}]{szegedy2015going}
Christian Szegedy, Wei Liu, Yangqing Jia, Pierre Sermanet, Scott Reed, Dragomir
  Anguelov, Dumitru Erhan, Vincent Vanhoucke, and Andrew Rabinovich.
\newblock Going deeper with convolutions.
\newblock In {\em CVPR}, 2015.

\bibitem[\protect\citeauthoryear{Vaswani \bgroup \em et al.\egroup
  }{2017}]{vaswani2017attention}
Ashish Vaswani, Noam Shazeer, Niki Parmar, Jakob Uszkoreit, Llion Jones,
  Aidan~N Gomez, {\L}ukasz Kaiser, and Illia Polosukhin.
\newblock Attention is all you need.
\newblock In {\em NeurIPS}, 2017.

\bibitem[\protect\citeauthoryear{Wu \bgroup \em et al.\egroup
  }{2015}]{wu2015object}
Y.~Wu, J.~Lim, and M.~Yang.
\newblock {Object tracking benchmark}.
\newblock {\em TPAMI}, 2015.

\bibitem[\protect\citeauthoryear{Yu \bgroup \em et al.\egroup
  }{2020}]{yu2020low}
Haibao Yu, Tuopu Wen, Guangliang Cheng, Jiankai Sun, Qi~Han, and Jianping Shi.
\newblock Low-bit quantization needs good distribution.
\newblock In {\em CVPRW}, 2020.

\bibitem[\protect\citeauthoryear{Zhu \bgroup \em et al.\egroup
  }{2018}]{zheng2018distractor}
Zheng Zhu, Qiang Wang, Bo~Li, Wei Wu, Junjie Yan, and Weiming Hu.
\newblock {Distractor-aware siamese networks for visual object tracking}.
\newblock In {\em ECCV}, 2018.

\bibitem[\protect\citeauthoryear{Zhuang \bgroup \em et al.\egroup
  }{2020}]{zhuang2020training}
Bohan Zhuang, Lingqiao Liu, Mingkui Tan, Chunhua Shen, and Ian Reid.
\newblock Training quantized neural networks with a full-precision auxiliary
  module.
\newblock In {\em CVPR}, 2020.

\end{thebibliography}

\end{document}